\documentclass[10pt,twocolumn,letterpaper]{article}

\usepackage[pagenumbers]{cvpr} 

\usepackage{amsmath,amsfonts,bm}

\def\eqref#1{equation~\ref{#1}}

\def\1{\bm{1}}

\DeclareMathAlphabet{\mathsfit}{\encodingdefault}{\sfdefault}{m}{sl}
\SetMathAlphabet{\mathsfit}{bold}{\encodingdefault}{\sfdefault}{bx}{n}

\newcommand{\R}{\mathbb{R}}

\DeclareMathOperator*{\argmin}{arg\,min}

\usepackage[dvipsnames]{xcolor}
\usepackage[pagebackref=true,breaklinks=true,letterpaper=true,colorlinks, bookmarks=false]{hyperref}
\usepackage{url}
\usepackage{graphicx}
\usepackage{booktabs}
\usepackage{multirow}
\usepackage{float}
\usepackage{wrapfig}
\usepackage{enumitem}
\usepackage{animate}
\usepackage[accsupp]{axessibility}  
\usepackage[capitalize]{cleveref}
\crefname{section}{Sec.}{Secs.}
\Crefname{section}{Section}{Sections}
\Crefname{table}{Table}{Tables}
\crefname{table}{Tab.}{Tabs.}

\def\cvprPaperID{1253} 
\def\confName{CVPR}
\def\confYear{2023}

\newcommand{\methodname}{BA-Det}
\newcommand\red[1]{\textcolor{red}{#1}}

\begin{document}

\title{3D Video Object Detection with Learnable Object-Centric Global Optimization}

\author{
        Jiawei He$^{1,2}$ \quad 
        Yuntao Chen$^{3}$ \quad 
        Naiyan Wang$^{4}$ \quad 
        Zhaoxiang Zhang$^{1,2,3}$ \\
        $^{1}$ CRIPAC, Institute of Automation, Chinese Academy of Sciences (CASIA)\\
        $^{2}$ School of Artificial Intelligence, University of Chinese Academy of Sciences (UCAS)\\
        $^{3}$ Centre for Artificial Intelligence and Robotics, HKISI\_CAS\quad
        $^{4}$ TuSimple\\
        {\tt\small
 \{hejiawei2019, zhaoxiang.zhang\}@ia.ac.cn \{chenyuntao08, winsty\}@gmail.com}
}

\maketitle

\begin{abstract}
We explore long-term temporal visual correspondence-based optimization for 3D video object detection in this work.
Visual correspondence refers to one-to-one mappings for pixels across multiple images.
Correspondence-based optimization is the cornerstone for 3D scene reconstruction but is less studied in 3D video object detection, because moving objects violate multi-view geometry constraints and are treated as outliers during scene reconstruction.
We address this issue by treating objects as first-class citizens during correspondence-based optimization.
In this work, we propose \methodname{}, an end-to-end optimizable object detector with object-centric temporal correspondence learning and featuremetric object bundle adjustment.
Empirically, we verify the effectiveness and efficiency of \methodname{} for multiple baseline 3D detectors under various setups.
Our \methodname{} achieves SOTA performance on the large-scale Waymo Open Dataset (WOD) with only marginal computation cost. 
Our code is available at \url{https://github.com/jiaweihe1996/BA-Det}.

\end{abstract}
\section{Introduction}
\begin{figure*}[t]
\centering
      \includegraphics[width = 0.88\textwidth]{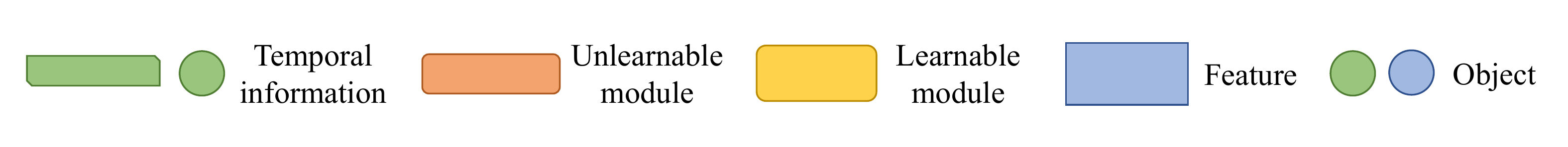}
      \vspace{-15pt}
\end{figure*}
\begin{figure*}[t]
\centering

	\subfloat[Temporal Filtering]{\includegraphics[width = 0.22\textwidth]{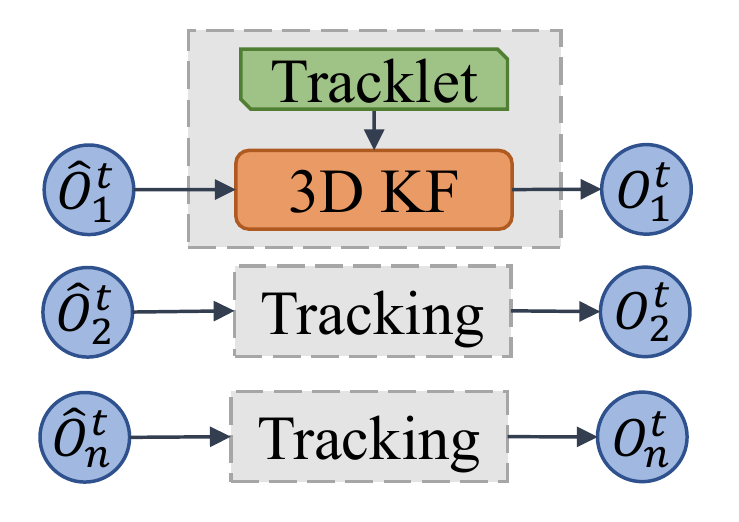}\label{fig1a}}
	\hfill
	\subfloat[Temporal BEV]{\includegraphics[width = 0.22\textwidth]{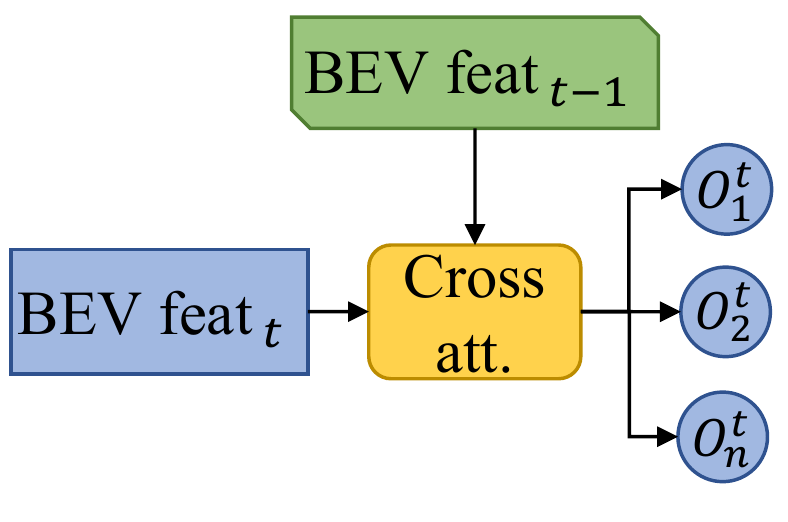}\label{fig1b}}
	\hfill
	\subfloat[Stereo from Video]{\includegraphics[width = 0.22\textwidth]{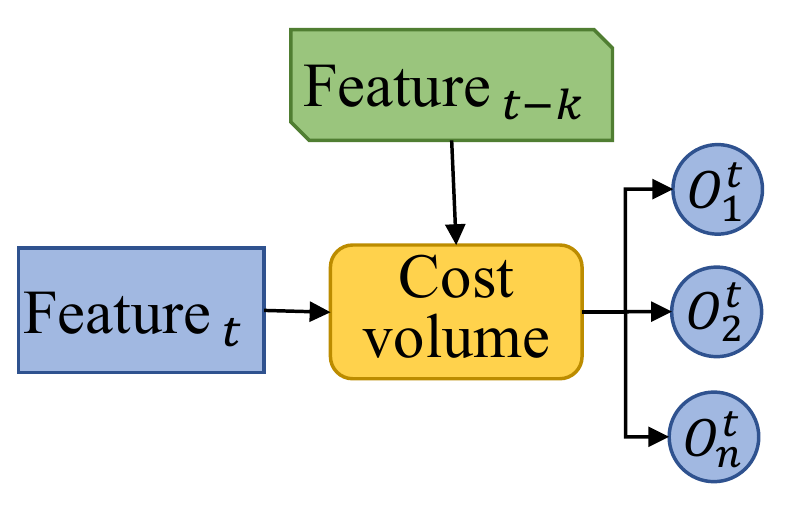}\label{fig1c}} 
        \hfill
        \subfloat[\methodname{} (Ours)]{\includegraphics[width = 0.22\textwidth]{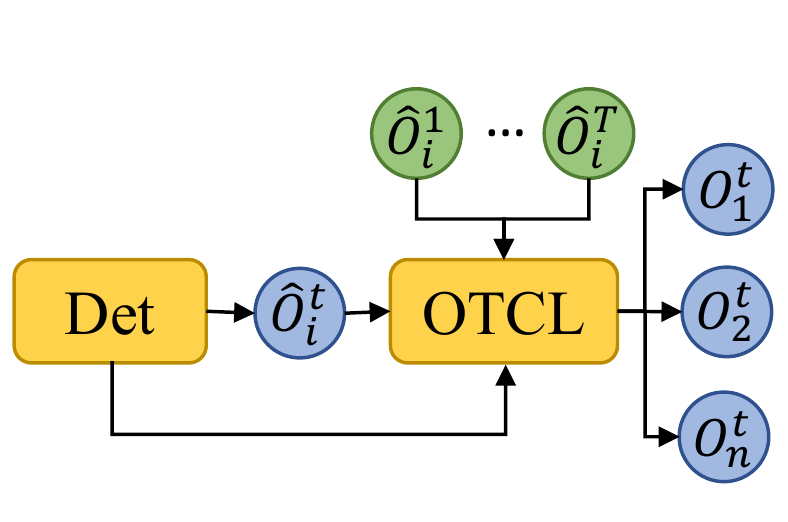}\label{fig1d}}
\caption{Illustration of how to leverage temporal information in different 3D video object detection paradigms.}
\label{fig1}
\end{figure*}
3D object detection is an important perception task, especially for indoor robots and autonomous-driving vehicles. 
Recently, image-only 3D object detection~\cite{zhang2021objects,li2022bevformer} has been proven practical and made great progress. 
In real-world applications, cameras capture video streams instead of unrelated frames, which suggests abundant temporal information is readily available for 3D object detection. 
In single-frame methods, despite simply relying on the prediction power of deep learning, 
finding correspondences play an important role in estimating per-pixel depth and the object pose in the camera frame. 
Popular correspondences include Perspective-n-Point (PnP) between pre-defined 3D keypoints ~\cite{zhang2021objects,li2022dcd} and their 2D projections in monocular 3D object detection, and Epipolar Geometry~\cite{chen2020dsgn,guo2021liga} in multi-view 3D object detection.
However, unlike the single-frame case, temporal visual correspondence has not been explored much in 3D video object detection.

As summarized in Fig.~\ref{fig1}, existing methods in 3D video object detection can be divided into three categories while each has its own limitations. 
Fig.~\ref{fig1a} shows methods with object tracking~\cite{brazil2020kinematic}, especially using a 3D Kalman Filter to smooth the trajectory of each detected object. 
This approach is detector-agnostic and thus widely adopted, but it is just an output-level smoothing process without any feature learning. As a result, the potential of video is under-exploited. 
Fig.~\ref{fig1b} illustrates the temporal BEV (Bird's-Eye View) approaches ~\cite{li2022bevformer,huang2022bevdet4d,liu2022petrv2} for 3D video object detection.
They introduce the multi-frame temporal cross-attention or concatenation for BEV features in an end-to-end fusion manner. 
As for utilizing temporal information, temporal BEV methods rely solely on feature fusion while ignoring explicit temporal correspondence.
Fig.~\ref{fig1c} depicts stereo-from-video methods~\cite{wang2022dfm,wang2022sts}.
These methods explicitly construct a pseudo-stereo view using ego-motion and then utilize the correspondence on the epipolar line of two frames for depth estimation.
However, the use of explicit correspondence in these methods is restricted to only two frames, thereby limiting its potential to utilize more temporal information.
Moreover, another inevitable defect of these methods is that moving objects break the epipolar constraints, which cannot be well handled, so monocular depth estimation has to be reused.



Considering the aforementioned shortcomings, we seek a new method that can \emph{handle both static and moving objects}, and \emph{utilize long-term temporal correspondences}. 
Firstly, in order to handle both static and moving objects, we draw experience from the object-centric global optimization with reprojection constraints in Simultaneous Localization and Mapping (SLAM)~\cite{yang2019cubeslam,li2018stereo}. 
Instead of directly estimating the depth for each pixel from temporal cues, we utilize them to construct useful temporal constraints to refine the object pose prediction from network prediction.
Specifically, we construct a non-linear least-square optimization problem with the temporal correspondence constraint in an object-centric manner to optimize the pose of objects no matter whether they are moving or not. 
Secondly, for long-term temporal correspondence learning, hand-crafted descriptors like SIFT~\cite{lowe2004sift} or ORB~\cite{rublee2011orb} are no longer suitable for our end-to-end object detector. Besides, the long-term temporal correspondence needs to be robust to viewpoint changes and severe occlusions, where these traditional sparse descriptors are incompetent. So, we expect to learn a dense temporal correspondence for all available frames.\par
In this paper, as shown in Fig.~\ref{fig1d}, we propose a 3D video object detection paradigm with learnable long-term temporal visual correspondence, called \emph{\methodname{}}. 
Specifically, the detector has two stages. In the first stage, a CenterNet-style monocular 3D object detector is applied for single-frame object detection. 
After associating the same objects in the video, the second stage detector extracts RoI features for the objects in the tracklet and matches dense local features on the object among multi-frames, called the object-centric temporal correspondence learning (OTCL) module.
To make traditional object bundle adjustment (OBA) learnable, we formulate featuremetric OBA.
In the training time, with featuremetric OBA loss, the object detection and temporal feature correspondence are learned jointly.
During inference, we use the 3D object estimation from the first stage as the initial pose and associate the objects with 3D Kalman Filter. The object-centric bundle adjustment refines the pose and 3D box size of the object in each frame at the tracklet level, taking the initial object pose and temporal feature correspondence from OTCL as the input. 
Experiment results on the large-scale Waymo Open Dataset (WOD) show that our \methodname{} could achieve state-of-the-art performance compared with other single-frame and multi-frame object detectors. 
We also conduct extensive ablation studies to demonstrate the effectiveness and efficiency of each component in our method.\par
In summary, our work has the following contributions:
\begin{itemize}[leftmargin=*]
    \item We present a novel object-centric 3D video object detection approach \emph{\methodname{}} by learning object detection and temporal correspondence jointly. 
   \item We design the second-stage object-centric temporal correspondence learning module and the featuremetric object bundle adjustment loss.
    \item We achieve state-of-the-art performance on the large-scale WOD. The ablation study and comparisons show the effectiveness and efficiency of our \methodname{}.
\end{itemize}

\section{Related Work}
\subsection{3D Video Object Detection}
For 3D video object detection, LiDAR-based methods~\cite{caesar2020nuscenes,yin2021center,fan2022embracing} usually align point clouds from consecutive frames by compensating ego-motion and simply accumulate them to alleviate the sparsity of point clouds. Object-level methods~\cite{qi2021offboard,you2022hindsight,chen2022mppnet,fan2023super}, handling the multi-frame point clouds of the tracked object, become a new trend. 
3D object detection from the monocular video has not received enough attention from researchers. Kinematic3D~\cite{brazil2020kinematic} is a pioneer work decomposing kinematic information into ego-motion and target object motion. However, they only apply 3D Kalman Filter~\cite{kalman1960new} based motion model for kinematic modeling and only consider the short-term temporal association (4 frames). Recently, BEVFormer~\cite{li2022bevformer} proposes an attentional transformer method to model the spatial and temporal relationship in the bird’s-eye-view (BEV). A concurrent work, DfM~\cite{wang2022dfm}, inspired by Multi-view Geometry, considers two frames as stereo and applies the cost volume in stereo to estimate depth. However, how to solve the moving objects is not well handled in this paradigm.

\subsection{Geometry in Videos}
Many researchers utilize 3D geometry in videos to reconstruct the scene and estimate the camera pose, which is a classic topic of computer vision. Structure from Motion (SfM)~\cite{schoenberger2016sfm} and Multi-view Stereo (MVS)~\cite{schoenberger2016mvs} are two paradigms to estimate the sparse and dense depth from multi-view images respectively. In robotics, 3D geometry theory is applied for Simultaneous Localization and Mapping (SLAM)~\cite{mur2015orb}. To globally optimize the 3D position of the feature points and the camera pose at each time, bundle adjustment algorithm~\cite{triggs1999bundle} is widely applied. However, most of them can only handle static regions in the scene. \par

In the deep learning era, with the development of object detection, object-level semantic SLAM~\cite{nicholson2018quadricslam,li2018stereo,yang2019cubeslam} is rising, aiming to reconstruct the objects instead of the whole scene. These methods can handle dynamic scenes and help the object localization in the video. Besides, feature correspondence learning~\cite{sarlin2020superglue,sun2021loftr} has received extensive attention in recent years. Deep learning has greatly changed the pipeline of feature matching. Differentiable bundle adjustment, like BANet~\cite{tang2018banet} and NRE~\cite{germain2021neural}, makes the whole 3D geometry system end-to-end learnable. Unlike these works, we focus on the representation of the 3D object and integrate feature correspondence learning into 3D object detection. Utilizing the learned temporal feature correspondence, the proposed \methodname{} optimizes the object pose of a tracklet in each frame.\par

\begin{figure*}[t]
          \centering
           \includegraphics[width=0.85\linewidth]{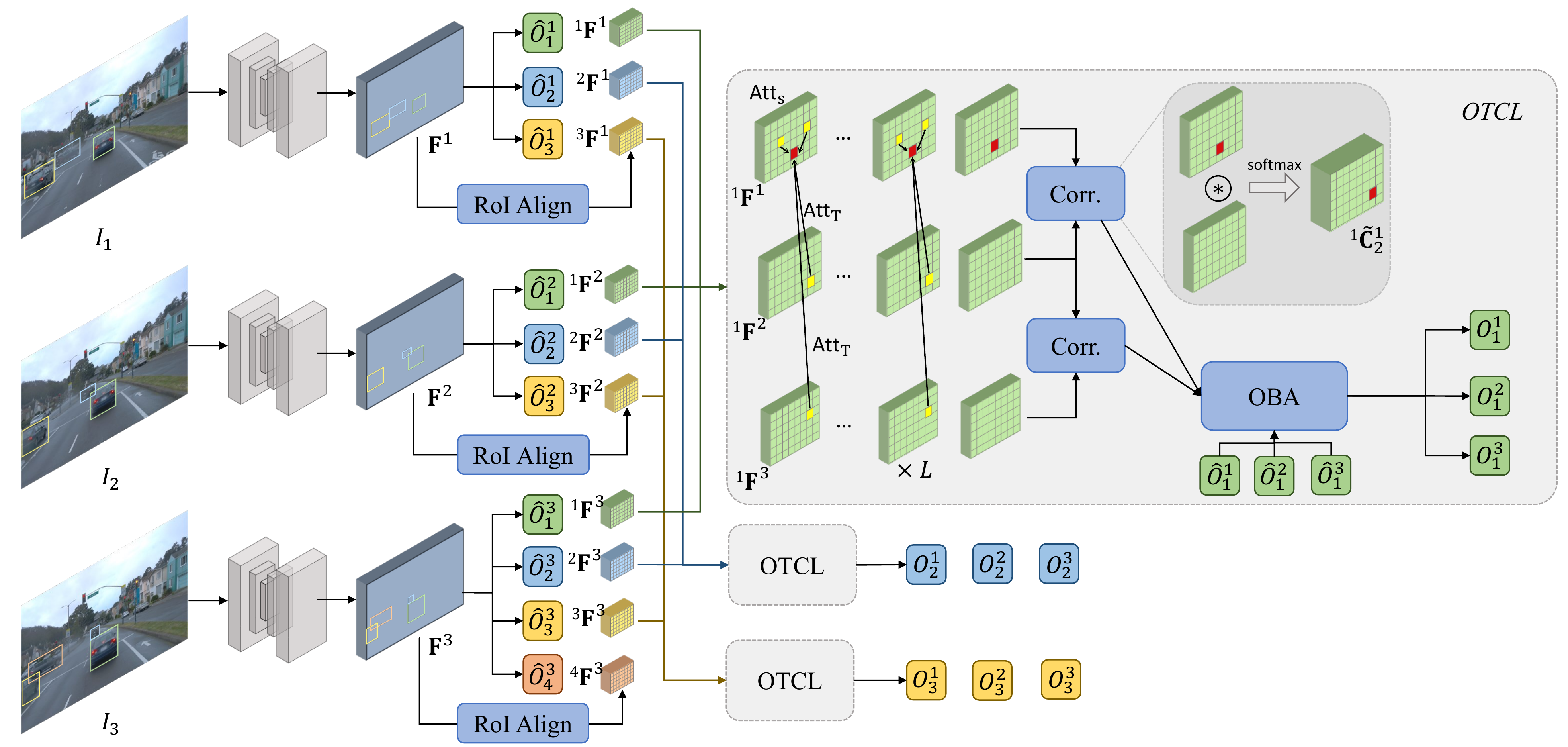}
             \caption{A overview of the proposed \methodname{} framework. The left part of the framework is the first-stage object detector to predict the 3D object and its 2D bounding box. The second stage is called \emph{OTCL} module. In the OTCL module, we extract the RoI features $^k\mathbf{F}^t$ by RoIAlign, aggregate the RoI features and learn object-centric temporal correspondence using featuremetric object bundle adjustment loss.}
             \label{fig:pipeline}
\end{figure*} 
\section{Preliminary: Bundle Adjustment}
Bundle Adjustment~\cite{triggs1999bundle} is a widely used globally temporal optimization technology in 3D reconstruction, which means optimally adjusting bundles of light rays from a given 3D global position to the camera center among multi-frames. 
Specifically, we use $\mathbf{P}_i=[x_i,y_i,z_i]^\top$ to denote the $i$-th 3D point coordinates in the global reference frame. 
According to the perspective camera model, the image coordinates of the projected 3D point at time $t$ is
\begin{equation}
\Pi(\mathbf{T}^t_{cg},\mathbf{P}_i,\mathbf{K})=\frac{1}{z_i^t}\mathbf{K}(\mathbf{R}^t_{cg}\mathbf{P}_i+\mathbf{t}^t_{cg}),
\end{equation}
where $\Pi$ is the perspective projection transformation, $\mathbf{T}^t_{cg}=[\mathbf{R}^t_{cg}|\mathbf{t}^t_{cg}]$ is the camera extrinsic matrix at time $t$.
$\mathbf{R}^t_{cg}$ and $\mathbf{t}^t_{cg}$ are the rotation and the translation components of $\mathbf{T}^t_{cg}$, respectively. 
$\mathbf{K}$ is the camera intrinsic matrix, and
$z_i^t$ is the depth of the $i$-th 3D point in the camera frame at time $t$.

Bundle adjustment is a nonlinear least-square problem to minimize the reprojection error as:
\begin{equation}
\begin{split}
    &\{\bar{\mathbf{T}}^t_{cg}\}_{t=1}^T,\{\bar{\mathbf{P}}_{i}\}_{i=1}^m=\\
    &\argmin_{\{\mathbf{T}^t_{cg}\}_{t=1}^T,\{\mathbf{P}_i\}_{i=1}^m}\frac{1}{2}\sum_{i=1}^{m}\sum_{t=1}^{T}||\mathbf{p}_i^t-\Pi(\mathbf{T}^t_{cg},\mathbf{P}_i,\mathbf{K})||^2,
\end{split}
\end{equation}
where $\mathbf{p}_i^t$ is the observed image coordinates of 3D point $\mathbf{P}_i$ on frame $t$.
Bundle adjustment can be solved by Gauss-Newton or Levenberg–Marquardt algorithm effectively~\cite{Agarwal_Ceres_Solver_2022,kummerle2011g2o}.

\section{\methodname{}: Object-centric Global Optimizable Detector}
\label{sec:det}
In this section, we introduce the framework of our \methodname{} (Fig.~\ref{fig:pipeline}), a learnable object-centric global optimization network. 
The pipeline consists of three parts: (1) First-stage single frame 3D object detection; (2) Second-stage object-centric temporal correspondence learning (OTCL) module; (3) Featuremetric object bundle adjustment loss for temporal feature correspondence learning.
\subsection{Single-frame 3D Object Detection}
Given a video clip with consecutive frames $\mathcal{V}=\{I_1,I_2,\cdots,I_T\}$, 3D video object detection is to predict the class and the 3D bounding box of each object in each frame. 
Let $\mathcal{O}_k^t$ be the $k$-th object in frame $t$. 
For the 3D bounding box $\mathbf{B}_k^t$, we estimate the size of the bounding box $\mathbf{s}_t^k=[w,h,l]^\top$ and the object pose $^k\mathbf{T}^{t}_{co}$ in the camera frame, including translation $^k\mathbf{t}^{t}_{co}=[x_c,y_c,z_c]^\top$ and rotation $^k\mathbf{r}^{t}_{co}=[r_x,r_y,r_z]^\top$. 
In most 3D object detection datasets, with the flat ground assumption, only yaw rotation $r_y$ is considered.
\par
We basically adopt MonoFlex~\cite{zhang2021objects} as our first-stage 3D object detector, which is a simple and widely-used baseline method.
Different from the standard MonoFlex, we make some modifications for simplicity and adaptation.
(1) Instead of ensemble the depth from keypoints and regression, we only used the regressed depth directly.
(2) The edge fusion module in MonoFlex is removed for simplicity and better performance.
The output of the first-stage object detector should be kept for the second stage. The predicted 2D bounding box $\mathbf{b}_k^t$ for each object is used for the object-centric feature extraction in the second stage. The 3D estimations should be the initial pose estimation and be associated between frames. We follow ImmortalTracker~\cite{wang2021immortal} to associate the 3D box prediction outputs with a 3D Kalman Filter frame by frame. For convenience and clarity, we use the same index $k$ to denote the objects belonging to the same tracklet in the video from now on.

\subsection{Object-Centric Temporal Correspondence Learning}
\label{sec:learn2dfeat}
Based on the predictions from the first-stage detector, we propose an object-centric temporal correspondence learning (\emph{OTCL}) module, which plays an indispensable role in the learnable optimization.
Specifically, the OTCL module is designed to learn the correspondence of the dense features for the same object among all available frames. 
Given a video $\{I_1,I_2,\cdots,I_T\}$ and image features $\{\mathbf{F}^1,\mathbf{F}^2,\cdots,\mathbf{F}^T\}$ from the backbone in the first stage, we extract the RoI features $^k\mathbf{F}^t\in \R^{H\times W\times C}$ of the object $\mathcal{O}^t_k$ by the RoIAlign operation~\cite{he2017mask},
\begin{equation}
    ^k\mathbf{F}^t=\mathtt{RoIAlign}(\mathbf{F}^t,\mathbf{b}_k^{t}).
\end{equation}

We apply $L$ layers of cross- and self-attention operations before calculating the correspondence map to aggregate and enhance the spatial and temporal information for RoI features. 
Note that the object tracklet is available with the aforementioned tracker, so the cross-attention is applied between the objects in different frames for the same tracklet.
For each layer of attention operations between two adjacent frames $t$ and $t'$:
\begin{equation}
    \begin{cases}
    ^k\widetilde{\mathbf{F}}^{t}=\mathtt{Att_S}(Q,K,V)=\mathtt{Att_S}(^k\hat{\mathbf{F}}^{t},{}^k\hat{\mathbf{F}}^{t},{}^k\hat{\mathbf{F}}^{t}),\\
    ^k\widetilde{\mathbf{F}}^{t'}=\mathtt{Att_S}(Q,K,V)=\mathtt{Att_S}(^k\hat{\mathbf{F}}^{t'},{}^k\hat{\mathbf{F}}^{t'},{}^k\hat{\mathbf{F}}^{t'}),\\
    ^k\hat{\mathbf{F}}^{t'}=\mathtt{Att_T}(Q,K,V)=\mathtt{Att_T}(^k\widetilde{\mathbf{F}}^{t'},{}^k\widetilde{\mathbf{F}}^{t},{}^k\widetilde{\mathbf{F}}^{t}),
    \end{cases}
\end{equation}
where $^k\hat{\mathbf{F}}^{t} \in \mathbb{R}^{HW\times C}$ is the flattened RoI feature, $\mathtt{Att_S}$ is the spatial self-attention, $\mathtt{Att_T}$ is the temporal cross-attention.\par
We then define the spatial correspondence map between two flattened RoI features after the attention operations. 
In frame pair $(t,t')$, we use $^k\mathbf{f}_i$ to denote $i$-th local feature in $^k\hat{\mathbf{F}}^{(L)}$ ($i\in \{1,2, \cdots, HW\}$).
The correspondence map $^k\mathbf{C}_{t}^{t'}\in \R^{HW\times HW}$ in two frames is defined as the inner product of two features in two frames:
\begin{equation}
    ^k\mathbf{C}_{t}^{t'}[i,i'] = {^k\mathbf{f}_i^{t}}*{^{k}\mathbf{f}_{i'}^{t'}}.
\end{equation}
To normalize the correspondence map, we perform softmax over all spatial locations $i'$,
\begin{equation}
    ^k\widetilde{\mathbf{C}}_{t}^{t'}[i,i'] = \mathtt{softmax}(^k\mathbf{C}_{t}^{t'}[i,i']).
\end{equation}

\subsection{Featuremetric Object Bundle Adjustment Loss}

In this subsection, we present that how to adapt and integrate the Object-centric Bundle Adjustment (OBA) into our learnable \methodname{} framework, based on the obtained correspondence map. Generally speaking, we formulate the featuremetric OBA loss to supervise the temporal feature correspondence learning. Note that here we only derive the tracklet-level OBA loss for the same object, and for the final supervision we will sum all the tracklet-level loss in the video.\par
\begin{figure}[t]
\centering
	\subfloat[Object-centric Bundle Adjustment (OBA).]{\includegraphics[width = 0.4\textwidth]{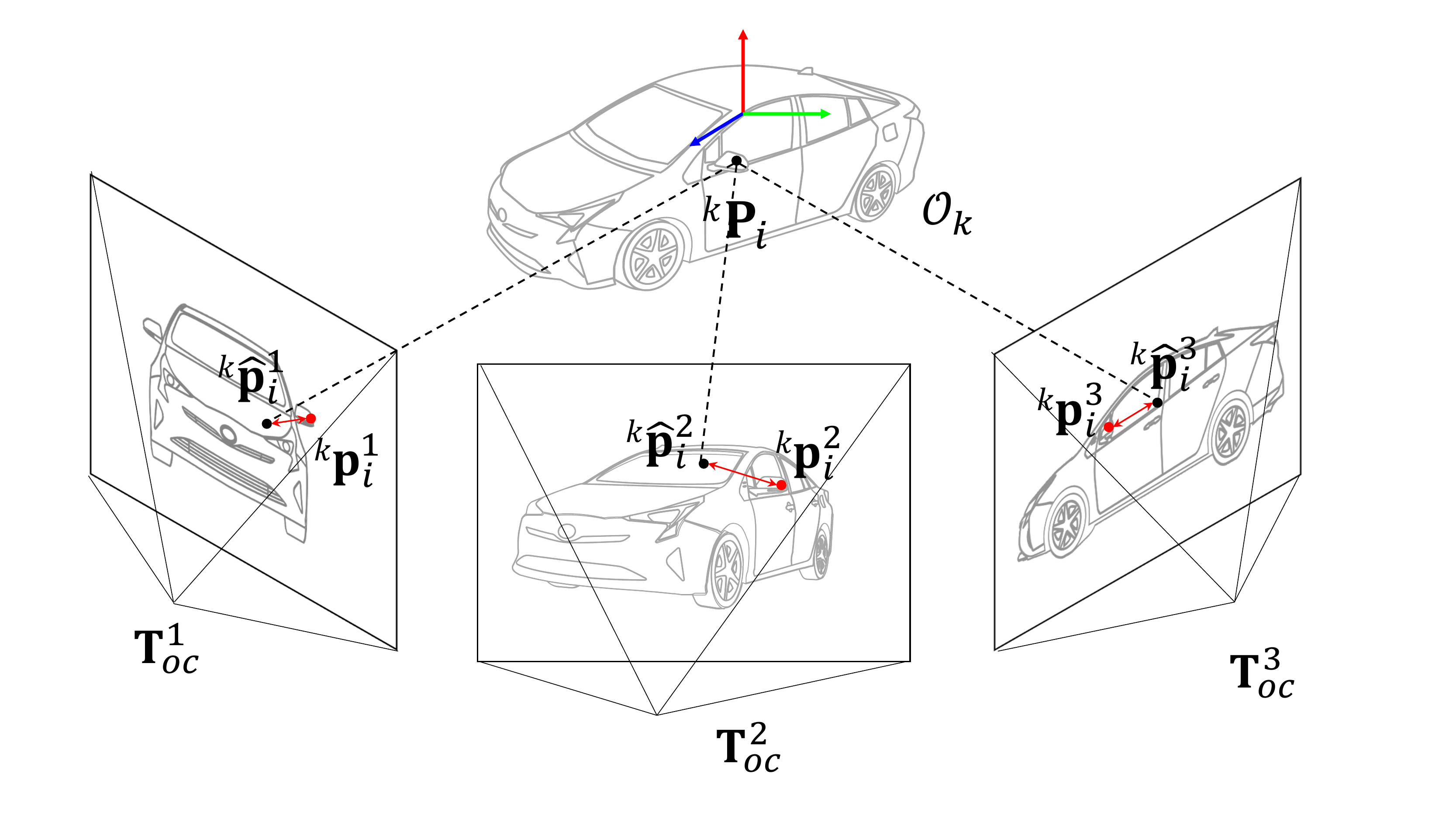}\label{fig:oba}}
	\hfill
        \subfloat[The computation of the featuremetric OBA loss.]{\includegraphics[width = 0.4\textwidth]{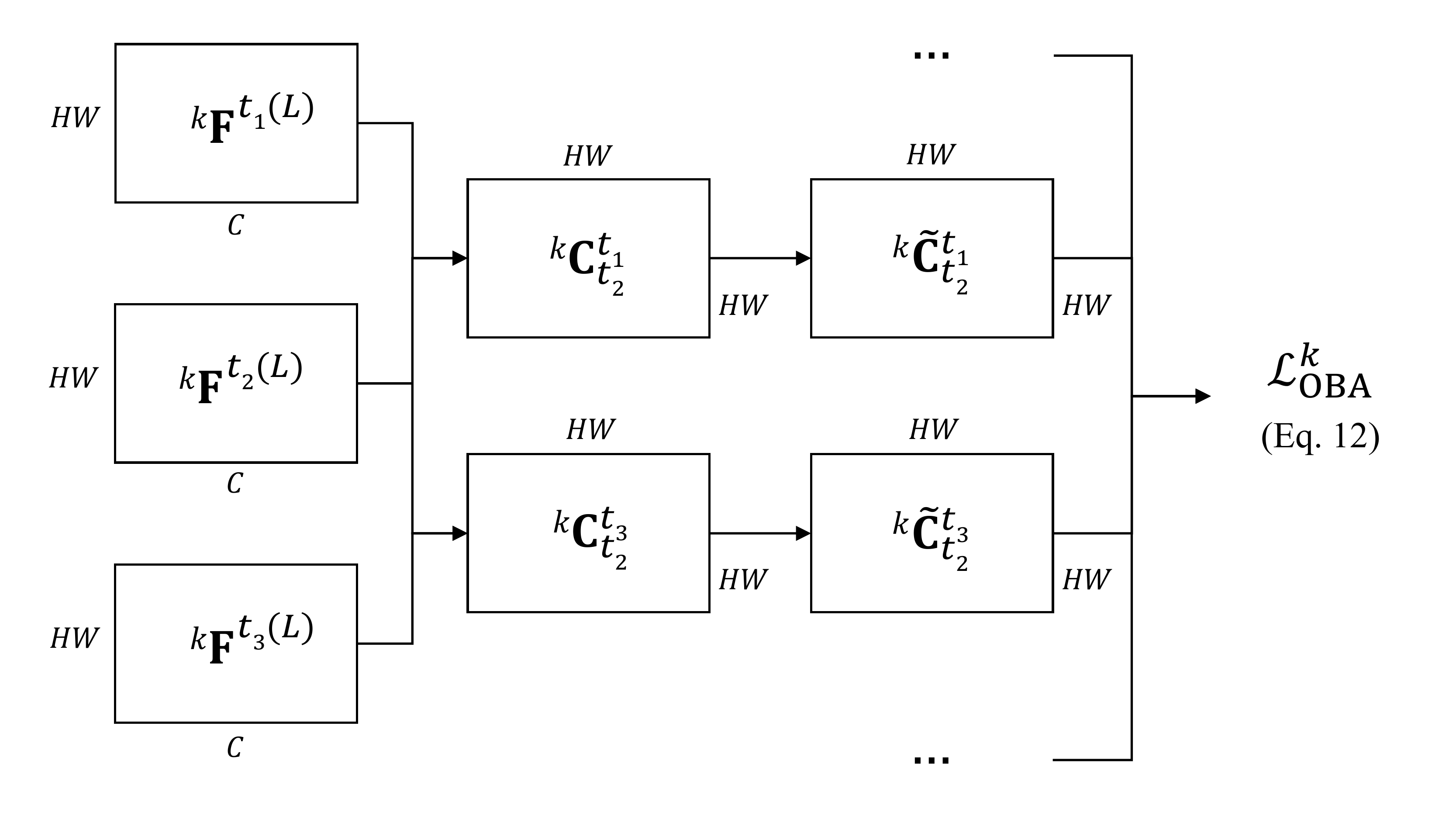}\label{fig:diffloss}}
\caption{Illustration of featuremetric object bundle adjustment.}
\vspace{-10pt}
\end{figure}

First, we revisit the object-centric bundle adjustment, as shown in Fig.~\ref{fig:oba}. 
As proposed in Object SLAM~\cite{yang2019cubeslam,li2018stereo}, OBA assumes that the object can only have rigid motion relative to the camera. 
For the object $\mathcal{O}_k$, we denote the 3D points as $\mathcal{P}_{k}=\{^{k}\mathbf{P}_i\}_{i=1}^m$ in the object frame, 2D points as $\{^k\mathbf{p}_i^t\}_{i=1}^m$, 2D features at position $^k\mathbf{p}_i^t$ as $\{\mathbf{f}[^k\mathbf{p}_i^t]\}_{i=1}^m$, and the camera pose in the object reference frame as $\mathcal{T}_k=\{^k\mathbf{T}^{t}_{co}\}_{t=1}^T$, OBA can be casted as:
\begin{equation}
\begin{split}
    &\bar{\mathcal{T}}_k,\bar{\mathcal{P}}_k=\argmin_{\mathcal{T}_k,\mathcal{P}_k}\frac{1}{2}\sum_{i=1}^{m}\sum_{t=1}^{T}||^k\mathbf{p}_i^t-\Pi(^k\mathbf{T}^{t}_{co},{^k\mathbf{P}_i},\mathbf{K})||_2^2.\\
\end{split}
\label{eq:objectba}
\end{equation}
To make the OBA layer end-to-end learnable, we formulate featuremetric~\cite{lindenberger2021pixel} OBA:
\begin{equation}
\begin{split}
    &\bar{\mathcal{T}}_k,\bar{\mathcal{P}}_k=\\
    &\argmin_{\mathcal{T}_k,\mathcal{P}_k}\frac{1}{2}\sum_{i=1}^{m}\sum_{t=1}^{T}\sum_{t'=1}^{T}||\mathbf{f}[^k\mathbf{p}_i^t]-\mathbf{f}[\Pi(^k\mathbf{T}^{t'}_{co},{^k\mathbf{P}_i},\mathbf{K})]||_2^2,
\end{split}
\label{eq:featobjectba}
\end{equation}
where $\mathbf{f}[\mathbf{p}]$ denotes the feature vector in pixel coordinates $\mathbf{p}$.
Representing the 3D point $^k\mathbf{P}_i$ in Eq.~\ref{eq:featobjectba} with 2D points in each frame, the featuremetric reprojection error of frame $t$ could be derived as
\begin{align}
    ^ke_{i}^{t}&=\sum_{t'=1}^T\mathbf{f}[{^k\mathbf{p}_i^{t}}]-\mathbf{f}[{^k\mathbf{p}_i^{t'}}]\\
    &=\sum_{t'=1}^T\mathbf{f}[{^k\mathbf{p}_i^{t}}]-\mathbf{f}[\Pi({^k{\mathbf{T}}^{t'}_{co}},\Pi^{-1}({^k{\mathbf{T}}^{t}_{co}},{}^k{\mathbf{p}}_i^{t},\mathbf{K}, {z}^{t}_i),\mathbf{K})],
\end{align}
where $\Pi^{-1}(\cdot)$ is the inverse projection function to lift the 2D point on the image to 3D in the object frame. ${z}_i^{t}$ is the ground-truth depth of $^k{\mathbf{p}}_i^{t}$ (from LiDAR point clouds only for training). In the training time, we learn the feature correspondence, given the ground-truth pose of the object $\mathcal{O}_k$, denoted as $^k{\mathbf{T}}^{{t}}_{co}$ and $^k{\mathbf{T}}^{{t'}}_{co}$ in frame $t$ and frame $t'$, respectively. 
Considering the featuremetric reprojection loss in all frames and all points,  the overall loss term for object $k$ can be formulated as
\begin{align}
    \mathcal{L}_{\text{rep}}^k&=\sum_{i=1}^m\sum_{t=1}^T||^ke_{i}^{t}||^2_2=\sum_{i=1}^m\sum_{t=1}^T\sum_{t'=1}^T||^k\mathbf{f}_i^{t}-{^k\mathbf{f}_i^{t'}}||^2_2
    \label{reploss}
\end{align}

Finally, we replace the $L2$ norm in Eq.~\ref{reploss} with the cosine distance to measure the featuremetric reprojection error.
Thus we bring the normalized correspondence map $\widetilde{\mathbf{C}}$ in Sec.~\ref{sec:learn2dfeat} into the loss term.
With log-likelihood formulation, we formulate the featuremetric OBA loss to supervise the object-centric temporal correspondence learning:
\begin{align}
\mathcal{L}_{\text{OBA}}^k
&=-\sum_{i=1}^m\sum_{t=1}^T\sum_{t'=1}^T\log(^k\widetilde{\mathbf{C}}_{t}^{t'}[{^k\bar{\mathbf{p}}_i^{t}},{^k\bar{\mathbf{p}}_i^{t'}}]).
\end{align}
where $({^k\bar{\mathbf{p}}_i^{t}},{^k\bar{\mathbf{p}}_i^{t'}})$ are the ground-truth corresponding pair of the $i$-th local feature. The illustration of the loss computation is in Fig.~\ref{fig:diffloss}.

\subsection{Inference}
\label{sec:infer}
After introducing the training loss design, we present the inference process of \methodname{} as follows. \par
\textbf{First-stage 3D object detection and association.} The first-stage detector makes the prediction of classification scores and 2D / 3D bounding boxes. The 3D bounding boxes are associated across the frames by ImmortalTracker~\cite{wang2021immortal}. The following process is on the tracklet level.\par
\textbf{Dense feature matching.} To optimize the object pose, we need to obtain the feature correspondence in each frame for the same object. As mentioned in Sec.~\ref{sec:learn2dfeat}, the OTCL module is trained to generate a dense correspondence map in all frames. During inference, we match  
 all $H\times W$ dense local features in RoI between adjacent two frames and between the first frame and last frame of the time window $[t,t+\tau]$. We use the RANSAC algorithm~\cite{fischler1981ransac} to filter the feature correspondence outliers. \par
\textbf{Feature tracking.}
To form a long-term keypoint tracklet from the obtained correspondence, we leverage a graph-based algorithm.
First, the matched feature pairs are constructed into a graph $\mathcal{G}$. The features are on the vertices. If the features are matched, an edge is connected in the graph. Then we track the feature for the object in all available frames. We use the association method mainly following~\cite{dusmanu2020multi}. The graph partitioning method is applied to $\mathcal{G}$ to make each connected subgraph have at most one vertex per frame. The graph cut is based on the similarity of the matched features.
\par
\textbf{Object-centric bundle adjustment.} In the inference stage, given the initial pose estimation and the temporal feature correspondence, we solve the object-centric bundle adjustment by Levenberg–Marquardt algorithm, and the object pose in each frame and the 3D position of the keypoints can be globally optimized between frames.\par
\textbf{Post-processing.} We also apply some common post-processing in video object detection techniques like tracklet rescoring~\cite{kang2017tcnn} and bounding box temporal interpolation.

\begin{table*}[]
\centering
\resizebox{0.9\linewidth}{!}{
\begin{tabular}{l|cccc|cccc}
\toprule
&\multicolumn{4}{c|}{LEVEL\_1} 
&\multicolumn{4}{c}{LEVEL\_2} \\
& 3D AP$_\text{70}$ & 3D APH$_\text{70}$ & 3D AP$_\text{50}$ & 3D APH$_\text{50}$ & 3D AP$_\text{70}$ & 3D APH$_\text{70}$ & 3D AP$_\text{50}$ & 3D APH$_\text{50}$ \\
\midrule
M3D-RPN~\cite{brazil2019m3d_rpn} &0.35&0.34&3.79&3.63&0.33&0.33&3.61&3.46\\
PatchNet~\cite{patchnet}&0.39&0.37&2.92&2.74&0.38&0.36&2.42&2.28\\
PCT~\cite{pct}&0.89&0.88&4.20&4.15&0.66&0.66&4.03&3.99\\
MonoJSG~\cite{lian2022monojsg}&0.97&0.95&5.65&5.47&0.91&0.89&5.34&5.17\\
GUPNet~\cite{lu2021gup}&2.28&2.27&10.02&9.94&2.14&2.12&9.39&9.31\\
DEVIANT~\cite{kumar2022deviant}&2.69&2.67&10.98&10.89&2.52&2.50&10.29&10.20 \\
CaDDN~\cite{reading2021categorical_caddn} &5.03&4.99&17.54&17.31&4.49&4.45&16.51&16.28\\
DID-M3D~\cite{peng2022did}&-&-&20.66&20.47&-&-&19.37&19.19\\
BEVFormer~\cite{li2022bevformer}$\dag$&- &7.70&-&30.80&-&6.90&-&27.70\\
DCD~\cite{li2022dcd} &12.57&12.50&33.44&33.24&11.78&11.72&31.43&31.25\\
\midrule
MonoFlex~\cite{zhang2021objects} (Baseline) &11.70&11.64&32.26&32.06&10.96&10.90&30.31&30.12\\
\textbf{\methodname{}(Ours)$\dag$}&\textbf{16.60}&\textbf{16.45}&\textbf{40.93}&\textbf{40.51}&\textbf{15.57}&\textbf{15.44}&\textbf{38.53}&\textbf{38.12}\\
\bottomrule
\end{tabular}}
\caption{The results on WODv1.2~\cite{sun2020scalability} \emph{val} set. AP$_{70}$ denotes AP with IoU threshold at 0.7. AP$_{50}$ denotes AP IoU@0.5.$\dag$ denotes the method utilizing temporal information. }
\label{tab:sota}
\end{table*}
\begin{figure*}[t]
\centering
        \resizebox{0.92\linewidth}{!}{
	\subfloat[Frame 8.]{\includegraphics[width = 0.2\linewidth]{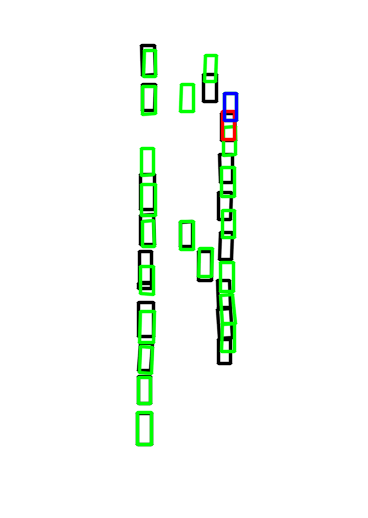}}
	\hfill
	\subfloat[Frame 22.]{\includegraphics[width = 0.2\linewidth]{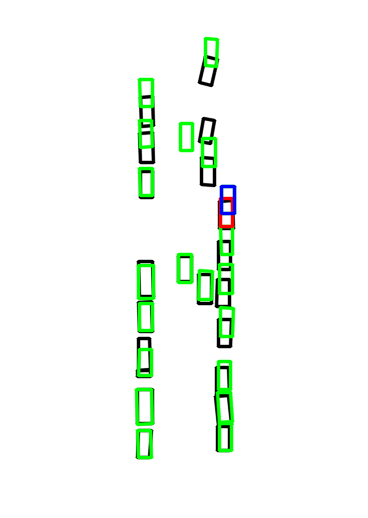}} 
        \hfill
        \subfloat[Frame 36.]{\includegraphics[width = 0.2\linewidth]{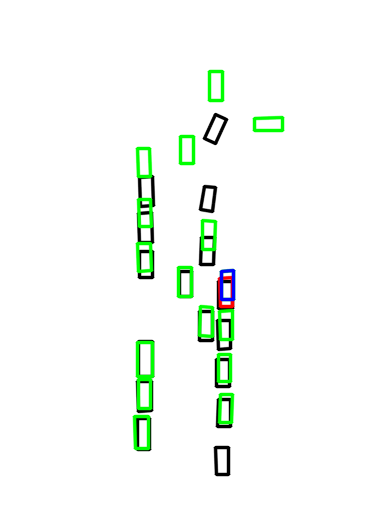}}
	\hfill
	\subfloat[Frame 50.]{\includegraphics[width = 0.2\linewidth]{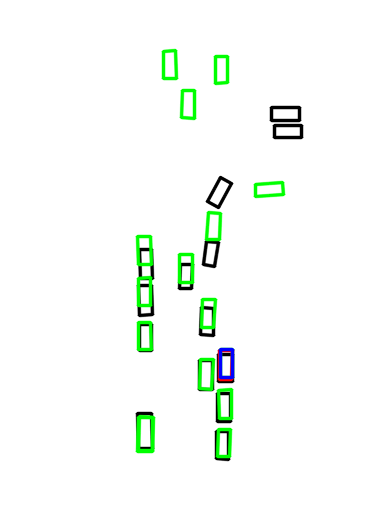}} 
        \hfill
        \subfloat[Frame 57.]{\includegraphics[width = 0.2\linewidth]{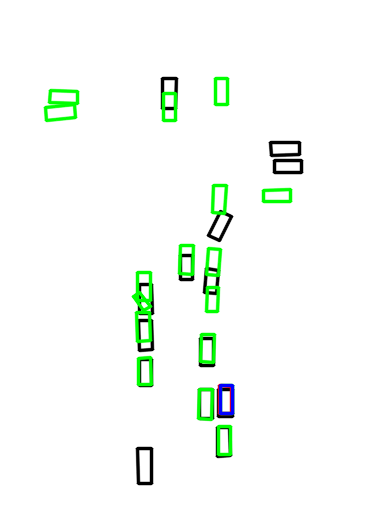}}
        }
\caption{Qualitative results from the BEV in different frames. We use \textcolor{blue}{blue} and \red{red} boxes to denote initial predictions and optimized predictions of the object we highlight. The \textcolor{green}{green} and black boxes denote the other box predictions and the ground truth boxes. The ego vehicle lies at the bottom of each figure.}
\label{fig:vismain}
\end{figure*}
\begin{table}[h]
\centering
\resizebox{\linewidth}{!}{
\begin{tabular}{l|ccc|cc}
\toprule
Method& LET-APL  & LET-AP & LET-APH & 3D AP$_{70}$ &3D AP$_{50}$\\
\midrule
MV-FCOS3D++~\cite{wang2022mvfcos3d++}$\dag$ &58.11 &74.68 &73.50 &14.66 &36.02 \\
\textbf{\methodname{}$_{\text{FCOS3D}}$(Ours)}$\dag$ &\textbf{58.47} &\textbf{74.85} &\textbf{73.66} &\textbf{15.02} &\textbf{36.89} \\
\bottomrule
\end{tabular}}
\caption{The multi-camera results on WODv1.3.1~\cite{hung2022let} \emph{val} set. Besides the official LET-IoU-based metrics, we also report the metrics with standard 3D IoU. All metrics are reported for the LEVEL\_2 difficulty.$\dag$: use temporal information. }
\label{tab:multi_cam}
\end{table}


\section{Experiments}
\subsection{Datasets and metrics}


\begin{table*}[]
\centering
\resizebox{0.92\linewidth}{!}{
\begin{tabular}{l|l|ccc|ccc|ccc|ccc}
\toprule
\multirow{2}{*}{} &\multirow{2}{*}{Method} &\multicolumn{3}{c|}{3D AP$_{70}$} &\multicolumn{3}{c|}{3D APH$_{70}$}&\multicolumn{3}{c|}{3D AP$_{50}$} &\multicolumn{3}{c}{3D APH$_{50}$} \\
 & &0-30 &30-50 &50-$\infty$ &0-30 &30-50 &50-$\infty$ &0-30 &30-50 &50-$\infty$ &0-30 &30-50 &50-$\infty$ \\ \midrule
  \multirow{3}{*}{L1} &DCD~\cite{li2022dcd} &32.47 &5.94 &1.24 &32.30 &5.91 &1.23 &62.70 &26.35 &10.16 &62.35 &26.21 &10.09 \\
 &MonoFlex~\cite{zhang2021objects} &30.64 &5.29 &1.05 &30.48 &5.27 &1.04&61.13 &25.85 &9.03 &60.75 &25.71 &8.95 \\
  &\textbf{\methodname{} (Ours)$\dag$}&\textbf{37.74}&\textbf{11.04}&\textbf{3.86}&\textbf{37.46}&\textbf{10.95}&\textbf{3.79}&\textbf{71.07}&\textbf{37.15}&\textbf{14.89}&\textbf{70.46}&\textbf{36.79}&\textbf{14.61}\\ 
\midrule
\multirow{3}{*}{L2} &
  DCD~\cite{li2022dcd} &32.30 &5.76 &1.08&32.19 &5.73 &1.08 &62.48 &25.60&8.92 &62.13 &25.46 &8.86\\
  &MonoFlex~\cite{zhang2021objects}&30.54 &5.14 &0.91 &30.37 &5.11 &0.91&60.91 &25.11 &7.92 &60.54 &24.97 &7.85 \\
 &\textbf{\methodname{} (Ours)$\dag$}&\textbf{37.61}&\textbf{10.72}&\textbf{3.37}&\textbf{37.33}&\textbf{10.63}&\textbf{3.31}&\textbf{70.83}&\textbf{36.14}&\textbf{13.62}&\textbf{70.23}&\textbf{35.79}&\textbf{13.37}\\ 
\bottomrule
\end{tabular}}
\caption{The object depth range conditioned result on WODv1.2~\cite{sun2020scalability} \emph{val} set. L1 and L2 denote LEVEL\_1 and LEVEL\_2 difficulty, respectively. $\dag$: use temporal information.}
\label{tab:range}
\end{table*}

We conduct our experiments on the large autonomous driving dataset, Waymo Open Dataset (WOD)~\cite{sun2020scalability}. The WOD has different versions with different annotations and metrics. To keep the fairness of the comparisons, we report the results both on WOD v1.2 and WOD v1.3.1. The annotations on v1.2 are based on LiDAR and the official metrics are mAP IoU@0.7 and mAP IoU@0.5. Recently, v1.3.1 is released to support multi-camera 3D object detection, and the annotations are camera-synced boxes. On the v1.3.1 dataset, a series of new LET-IoU-based metrics~\cite{hung2022let} are introduced to slightly tolerate the localization error from the worse sensor, camera, than LiDAR. Early work mainly reports the results on the v1.2 dataset, and we only compare our methods with the ones from WOD Challenge 2022 using the v1.3.1 dataset. Because we mainly focus on rigid objects, we report the results of the VEHICLE class.\par
LET-3D-AP and LET-3D-APL are the new metrics, relying on the Longitudinal Error Tolerant IoU (LET-IoU).  LET-IoU is the 3D IoU calculated between the target ground truth box and the prediction box aligned with ground truth along the depth that has minimum depth error.
LET-3D-AP and LET-3D-APL are calculated from the average precision and the longitudinal affinity weighted average precision of the PR curve.
For more details, please refer to \cite{hung2022let}.

\begin{table*}[]
\centering
\resizebox{0.98\linewidth}{!}{
\begin{tabular}{l|cccc|cccc}
\toprule
&\multicolumn{4}{c|}{LEVEL\_1} 
&\multicolumn{4}{c}{LEVEL\_2} \\
& 3D AP$_\text{70}$ & 3D APH$_\text{70}$ & 3D AP$_\text{50}$ & 3D APH$_\text{50}$ & 3D AP$_\text{70}$ & 3D APH$_\text{70}$ & 3D AP$_\text{50}$ & 3D APH$_\text{50}$ \\
\midrule
MonoFlex (baseline) &11.70&11.64&32.26&32.06&10.96&10.90&30.31&30.12\\
\midrule
Our first-stage prediction&13.57&13.48&34.70&34.43 &12.72 &12.64&32.56&32.32\\
+3D Tracking~\cite{wang2021immortal} &14.01 &13.93&35.19 &34.92 &13.13 &13.05 &33.03 &32.78\\
+ Learnable global optimization &15.85 &15.75 &38.06 &37.76 &14.87 &14.77 &35.72 &35.44\\
+ Tracklet rescoring &16.43 &16.30 &40.07 &39.70 &15.41&15.29&37.66 &37.31\\
+ Bbox interpolation &16.60&16.45&40.93&40.51&15.57&15.44&38.53&38.12\\
\bottomrule
\end{tabular}}
\caption{Ablation study of each component in \methodname{}.}
\label{tab:ablation}
\end{table*}

\begin{table*}[]
\centering
\resizebox{0.9\linewidth}{!}{
\begin{tabular}{l|cccc|cccc}
\toprule
&\multicolumn{4}{c|}{LEVEL\_1} 
&\multicolumn{4}{c}{LEVEL\_2} \\
& 3D AP$_\text{70}$ & 3D APH$_\text{70}$ & 3D AP$_\text{50}$ & 3D APH$_\text{50}$ & 3D AP$_\text{70}$ & 3D APH$_\text{70}$ & 3D AP$_\text{50}$ & 3D APH$_\text{50}$ \\
\midrule
MonoFlex (baseline) &11.70&11.64&32.26&32.06&10.96&10.90&30.31&30.12\\
Initial prediction&13.57&13.48&34.70&34.43 &12.72 &12.64&32.56&32.32\\
\midrule
Static BA  &14.73 &14.62 &37.89 &37.56 &13.82 &13.72 &35.65 &35.34\\
Ours   &16.60&16.45&40.93&40.51&15.57&15.44&38.53&38.12\\
\bottomrule
\end{tabular}}
\caption{Comparison between object-centric \methodname{} and the traditional scene-level bundle adjustment (Static BA). Initial prediction denotes the predictions in the first stage.}
\vspace{-5pt}
\label{tab:static}
\end{table*}

\subsection{Implementation Details}
\label{sec:impl}
The first stage network architecture of \methodname{} is the same as MonoFlex, with DLA-34~\cite{yu2018dla} backbone, the output feature map is with the stride of 8. In the second stage, the shape of the RoI feature is $60\times 80$. The spatial and temporal attention module is stacked with 4 layers. The implementation is based on the PyTorch framework. We train our model on 8 NVIDIA RTX 3090 GPUs for 14 epochs. Adam optimizer is applied with $\beta_1 =0.9$ and $\beta_2=0.999$. The initial learning rate is $5 \times 10^{-4}$ and weight decay is $10^{-5}$. The learning rate scheduler is one-cycle. We use the Levenberg-Marquardt algorithm, implemented by DeepLM~\cite{huang2021deeplm}, to solve object-centric bundle adjustment. The maximum iteration of the LM algorithm is 200. For the object that appears less than 10 frames or the average keypoint number is less than 5, we do not optimize it. 

\begin{table*}[]
\centering
\resizebox{0.98\linewidth}{!}{
\begin{tabular}{l|c|cccc|cccc}
\toprule
&  \multirow{2}{*}{$\Bar{L}_t$}
&\multicolumn{4}{c|}{LEVEL\_1} 
&\multicolumn{4}{c}{LEVEL\_2} \\
&& 3D AP$_\text{70}$ & 3D APH$_\text{70}$ & 3D AP$_\text{50}$ & 3D APH$_\text{50}$ & 3D AP$_\text{70}$ & 3D APH$_\text{70}$ & 3D AP$_\text{50}$ & 3D APH$_\text{50}$ \\
\midrule

MonoFlex (baseline) &-&11.70&11.64&32.26&32.06&10.96&10.90&30.31&30.12\\
\midrule
 \methodname{}+ ORB feature~\cite{rublee2011orb} &2.6&14.05&13.96&35.21&34.95&13.17&13.08&33.05&32.81\\
 \methodname{}+ Our feature &10 &16.60&16.45&40.93&40.51&15.57&15.44&38.53&38.12\\
\bottomrule
\end{tabular}}
\caption{Ablation study about different feature corresponding methods. $\Bar{L}_t$ denotes the average keypoint tracklet length for each object.}
\vspace{-10pt}
\label{tab:orb}
\end{table*}

\subsection{Comparisons with State-of-the-art Methods}
We compare our \methodname{} with other state-of-the-art methods under two different settings. WODv1.2 is for the front view camera and WODv1.3.1 has the official evaluator for all 5 cameras. As shown in Table~\ref{tab:sota}, using the FRONT camera, we outperform the SOTA method DCD~\cite{li2022dcd} for about 4AP and 4APH ($\sim$30\% improvement) under the 0.7 IoU threshold. Compared with the only temporal method BEVFormer~\cite{li2022bevformer}, we have double points of 3D AP$_{70}$ and 3D APH$_{70}$. To validate the effectiveness, we also report the multi-camera results on the newly released WODv1.3.1, as shown in Table~\ref{tab:multi_cam}. No published work reports the results on WODv1.3.1. So, we only compare with the open-source MV-FCOS3D++~\cite{wang2022mvfcos3d++}, the second-place winner of WOD 2022 challenge. We design the variant of \methodname{}, called \methodname{}$_\text{FCOS3D}$, to adapt to the multi-camera setting. \methodname{}$_\text{FCOS3D}$ is also a two-stage object detector. The first stage is the same as MV-FCOS3D++, but with the output of 2D bounding boxes. The second stage is OTCL module supervised with featuremetric object bundle adjustment loss. Although there are overlaps between 5 cameras, to simplify the framework, we ignore the object BA optimization across cameras and only conduct temporal optimization. \methodname{}$_\text{FCOS3D}$ outperforms MV-FCOS3D++ under main metrics and traditional 3D IoU-based metrics.
\subsection{Qualitative Results}
In Fig.~\ref{fig:vismain}, we show the object-level qualitative results of the first-stage and second-stage predictions in different frames. For a tracklet, we can refine the bounding box predictions with the help of better measurements in other frames, even if there is a long time interval between them.
\subsection{Distance Conditioned Results}
We report the results with the different depth ranges in Table~\ref{tab:range}. The results indicate that the single frame methods, like DCD and MonoFlex, are seriously affected by object depth. When the object is farther away from the ego vehicle, the detection performance drops sharply. Compared with these methods, \methodname{}, has the gain almost from the object far away from the ego-vehicle. The 3D AP$_{70}$ and 3D APH$_{70}$ are 3$\times$ compared with the baseline when the object is located in $[50\text{m},\infty)$, 2$\times$ in $[30\text{m},50\text{m})$ and 1.2$\times$ in $[0\text{m},30\text{m})$. This is because we utilize the long-term temporal information for each object. In a tracklet, the predictions near the ego-vehicle can help to refine the object far away.
\subsection{Ablation study}

We ablate each component of \methodname{}. The results are shown in Table~\ref{tab:ablation}. The first stage detector is slightly better than the MonoFlex baseline mainly because we remove the edge fusion module, which is harmful to the truncated objects in WOD. 3D KF associates the objects and smooths the object's trajectory. This part of improvement can be regarded as similar to Kinematic3D~\cite{brazil2020kinematic}. The core of \methodname{} is the learnable global optimization module, which obtains the largest gain in all modules. The tracklet rescoring and temporal interpolation modules are also useful.
\subsection{Further Discussions}

\noindent{\textbf{BA vs. Object BA.}} We conduct experiments to discuss whether the object-centric manner is important in temporal optimization. We modify our pipeline and optimize the whole scene in the \emph{global} frame instead of optimizing the object pose in the object frame, called Static BA in Table~\ref{tab:static}. Static BA ignores dynamic objects and treats them the same as static objects. The inability to handle dynamic objects causes decreases by about 2 AP compared with \methodname{}.\par
\noindent{\textbf{Temporal feature correspondence.}} As shown in Table~\ref{tab:orb}, we ablate the features used for object-centric bundle adjustment. Compared with traditional ORB feature~\cite{rublee2011orb}, widely used in SLAM, our feature learning module predicts denser and better correspondence. We find the average object tracklet length is 19.6 frames, and the average feature tracklet in our method is about 10 frames, which means we can keep a long feature dependency and better utilize long-range temporal information. However, the $\Bar{L}_t$ of the ORB feature is only 2.6 frames. The results show the short keypoint tracklet can not refine the long-term object pose well.\par

\noindent{\textbf{Inference latency of each step in \methodname{}.}} The inference latency of each step in \methodname{} is shown in Table~\ref{tab:latency}. The most time-consuming part is the first-stage object detector, more than 130ms per image, which is the same as the MonoFlex baseline. 
Our \methodname{} only takes an additional 50ms latency per image, compared with the single-frame detector MonoFlex. Besides, although the dense feature correspondence is calculated, thanks to the shared backbone with the first stage detector and parallel processing for the objects, the feature correspondence module is not very time-consuming.\par
\begin{table}[h]
\centering
\resizebox{0.6\linewidth}{!}{
\begin{tabular}{l|r}
\toprule
Total latency &181.5ms\\
\midrule
First-stage detector&132.6ms\\
Object tracking&6.6ms\\
Feature correspondence&23.0ms\\
Object bundle adjustment&19.3ms\\
\bottomrule
\end{tabular}}
\caption{Inference latency of each step in \methodname{} per image.}
\vspace{-10pt}
\label{tab:latency}
\end{table}

\section{Limitations and Future Work}
In the current version of this paper, we only focus on the objects, such as cars, trucks, and trailers. The performance of non-rigid objects such as pedestrians has not been investigated. However, with mesh-based and skeleton-based 3D human models, we believe that a unified keypoint temporal alignment module can be designed in the future. So, we will explore the extension of \methodname{} for non-rigid objects.
\section{Conclusion}
In this paper, we propose a 3D video object detection paradigm with long-term temporal visual correspondence, called BA-Det. \methodname{} is a two-stage object detector that can jointly learn object detection and temporal feature correspondence with proposed Featuremetric OBA loss. 
Object-centric bundle adjustment optimizes the first-stage object estimation globally in each frame. \methodname{} achieves state-of-the-art performance on WOD. \par

\section*{Acknowledgements} 
This work was supported in part by the Major Project for New Generation of AI (No.2018AAA0100400), the National Natural Science Foundation of China (No. 61836014, No. U21B2042, No. 62072457, No. 62006231) and the InnoHK program. The authors thank Lue Fan and Yuqi Wang for their valuable suggestions.
{\small
\bibliographystyle{ieee_fullname}
\bibliography{egbib}
}
\newpage
\section*{Appendix}
\renewcommand\thesection{\Alph{section}}
\renewcommand\thetable{\Alph{table}}
\renewcommand\thefigure{\Alph{figure}}
\setcounter{section}{0}
\setcounter{table}{0}
\setcounter{figure}{0}
\section{Network Architecture}
\label{sec:arc}
In this section, we explain the detailed network design of \methodname{}. The backbone is a standard `DLASeg' architecture, please refer to~\cite{yu2018dla} for more details. Then two detection heads are added in the first-stage detector. The cls head contains a 3$\times$3 Conv2d layer, and an FC to predict the classification. The reg head contains 8 independent `3$\times$3 Conv2d layer + FC' modules, regressing `2d dim', `3d offset', `corner offset', `corner uncertainty',`3d dim', `ori cls', `ori offset', `depth', `depth uncertainty' attributes. The `ori cls' and `ori offset' share the Conv2d layer. Note that the `inplace-abn' module in MonoFlex~\cite{zhang2021objects} is changed back to a BatchNorm2d layer and a ReLU activation function. The architecture of OTCL module includes RoIAlign, conv layer, attention layer, and correlation layer. We adopt torchvision's RoIAlign implementation\footnote{\url{https://github.com/pytorch/vision/blob/main/torchvision/ops/roi\_align.py}}. The output size of the RoI feature is 60$\times$80. The conv layer includes two `3$\times$3 Conv2d + BatchNorm2d + ReLU' modules. The attention layer contains 4 self-attention modules and 4 cross-attention modules, using standard MultiheadAttention implementation. The channel dimension is 64, and the number of attention heads is 4. The correlation layer contains the operation of the feature inner product and the softmax normalization.
\begin{table}[h]
\centering
\resizebox{0.98\linewidth}{!}{
\begin{tabular}{l|c}
\toprule
config & value\\
\midrule
optimizer & Adam\\
optimizer momentum &$\beta_1=0.9,\beta_2=0.999$ \\
weight decay & $1\times 10^{-5}$\\
learning rate &$5\times 10^{-4}$\\
learning rate schedule &cosine decay \\
warmup iterations &500 \\
epochs &14 \\
augmentation &random flip \\
aug prob &0.5 \\
batch size &8 \\
gradient clip &10 \\
image size & 1920 $\times$1280\\
\multirow{4}{*}{loss name} & heatmap loss, bbox loss, depth loss\\
&offset loss, orientation loss, dims loss\\
&corner loss, keypoint loss, kp depth loss\\
&trunc offset loss, featuremetric OBA loss\\
\multirow{4}{*}{loss weight} & 1.0        , 1.0           , 1.0   \\       
& 0.6  , 1.0            , 0.33      \\
& 0.025        , 0.01  ,0.066       \\
&               0.6                ,1.0 \\
sync BN & True\\

\bottomrule
\end{tabular}}
\caption{Training config of \methodname{}.}
\label{tab:config}
\end{table}
\begin{table}[h]
\centering
\resizebox{0.98\linewidth}{!}{
\begin{tabular}{l|c}
\toprule
config & value\\
\midrule
optimizer & AdamW\\
optimizer momentum &$\beta_1=0.9,\beta_2=0.999$ \\
weight decay & $1\times 10^{-4}$\\
learning rate &$2.5\times 10^{-4}$\\
learning rate schedule &step decay \\
lr decay epoch &[16, 22] \\
warmup iterations &500 \\
epochs &24 \\
data interval & 5\\
augmentation &random flip \\
aug prob &0.5 \\
batch size &8 \\
gradient clip &35 \\
image size & 1248 $\times$ 832\\
loss name & cls loss, bbox loss,dir loss, OBA loss\\
loss weight &1.0,2.0,0.2,0.8\\
gt assigner &MaxIoUAssigner\\
sync BN & True\\
pre-train & FCOS3D++\_r101\_DCN\_2x\_D3\\
\bottomrule
\end{tabular}}
\caption{Training config of \methodname{}$_{\text{FCOS3D}}$.}
\label{tab:config2}
\end{table}
\section{Training and Inference Details}
\label{sec:detail}
\subsection{Training Settings}
In Table~\ref{tab:config} and Table~\ref{tab:config2}, we show the training details of \methodname{} and the variant \methodname{}$_{\text{FCOS3D}}$.\par

\begin{figure*}[h]
          \centering
            \vspace{-30pt}
          \subfloat[segment-15948509588157321530\_7187\_290\_7207\_290]{\animategraphics[width=0.5\linewidth,autoplay=True,loop]{5}{./fig/vis/1075}{000}{074}}
	   \hfill
	   \subfloat[segment-16229547658178627464\_380\_000\_400\_000]{\animategraphics[width=0.5\linewidth,autoplay=True,loop]{5}{./fig/vis2/1079}{008}{066}} 
             \caption{Qualitative results in the WOD \emph{val} set as videos. We use \textcolor{blue}{blue} and \red{red} boxes to denote initial predictions and optimized predictions of the object we highlight. The \textcolor{green}{green} and \textbf{black} boxes denote the other box predictions and the ground truth boxes. The ego vehicle lies at the bottom of each figure. Please view with Adobe Acrobat Reader to see the videos.}
             \label{fig:vis}
\end{figure*} 
\begin{figure*}[h]
          \centering
\animategraphics[width=0.9\linewidth,autoplay=True,loop]{1}{./vis/}{000}{065}
\caption{Video with detection and tracking results. Open with Adobe Acrobat Reader to play the video. Best viewed by zooming in.} 
\vspace{-8pt}
\label{video}
\end{figure*}
\subsection{Inference Details}
In the main paper, we introduce the inference process in Sec.~\ref{sec:infer}. Here, we make some additional explanations of inference details.\par
\noindent{\bf{Dense feature matching.}} We match the dense RoI feature in two frames. The sliding window $\tau$ is 5 in the implementation. We adopt OPENCV's implementation of the RANSAC operation. To balance the number of valid features in each frame, we select the top $k$ feature correspondence for each frame based on the similarities. $k$ is set to 50 in the implementation.
 Note that if the number of correspondences is imbalanced between frames, the optimization tends to give high weight excessively to the frames with more tracklets and make other frames deviate from the correct pose.\par
\noindent{\bf{Object-centric bundle adjustment.}} We use DeepLM to solve the non-linear least-square optimization problem. Note that in the inference stage, we adopt the original OBA formulation with reprojection error, according to the correspondence prediction from the OTCL module. The initial 3D position of the keypoint is set to (0,0,0) in the object reference frame. In DeepLM, no robust loss (e.g., Huber loss) is used.\par
\noindent{\bf{Post-processing.}} For the tracklet rescoring process, we adopt the maximum predicted score of the tracklet to replace the first-stage predicted score in each frame. The bounding box interpolation is for the missing detection in the tracklet. This process is to interpolate the 3D location of the object center, 3D box size, and orientation in the global reference frame. Only the nearest two observations are considered in interpolation.\par
\section{Qualitative Results}
We show some additional qualitative results in the appendix. In Fig.~\ref{fig:vis}, the qualitative results of two sequences (segment-15948509588157321530\_7187\_290\_7207\_290 and segment-16229547658178627464\_380\_000\_400\_000) in the WOD~\cite{sun2020scalability} \emph{val} set are shown as videos. For better understanding, we show a more detailed video (Fig.~\ref{video}) with detection and tracking results in BEV. Please view with Adobe Acrobat to see the videos.
\section{Additional Experiments}
\begin{table}[t]
\centering
\resizebox{\linewidth}{!}{
\begin{tabular}{l|ccccc}
\toprule
 &MOTA&MOTP&Miss &Mismatch&FP\\
\midrule
1st + ImmortalTracker &0.190&0.282 &0.724&9.84e-04&0.085\\
BA-Det (Ours) &0.238&0.282&0.652 &5.44e-04&0.110\\
\bottomrule
\end{tabular}}
\caption{Tracking results evaluated on WOD val set. We report the metrics on LEVEL 2 @IoU0.5.}
\vspace{-10pt}
\label{tab:track}
\end{table}
\begin{table}[h]
\centering
\resizebox{0.7\linewidth}{!}{
\begin{tabular}{l|cc}
\toprule
Tracking methods &3D AP$_{70}$ &3D AP$_{50}$\\
\midrule
AB3DMOT~\cite{weng2020ab3dmot}$^*$ &16.3 &38.9\\
ImmortalTracker~\cite{wang2021immortal} &16.6&40.9\\
\bottomrule
\end{tabular}}
\caption{Influence of tracking methods. $^*$: implemented by us.}
\vspace{-10pt}
\label{tab:ab3d}
\end{table}
We evaluate the tracking results compared with the baseline of first-stage predictions + ImmortalTracker in Table~\ref{tab:track}.
Besides, to show how tracking quality affects detection, we use a weaker tracker, AB3DMOT~\cite{weng2020ab3dmot}. The results are shown in Table~\ref{tab:ab3d}. The results reveal that our BA-Det improves tracking performance and is robust to tracking quality.
\section{Reproducibility Statement}
We will release the training and inference codes to help reproducing our work. Limited by the license of WOD, the checkpoint of the model cannot be publicly available. However, we will provide the checkpoint by email. The implementation details, including the training and inference settings, network architecture, and the \methodname{}$_{\text{FCOS3D}}$ variant, are mentioned in Sec.~\ref{sec:det} and Sec.~\ref{sec:impl} in the main paper and Sec.~\ref{sec:arc} and \ref{sec:detail} in the appendix. WOD is a publicly available dataset.

\end{document}